\pdfoutput=1

\documentclass[11pt]{article}

\usepackage[final]{acl}

\usepackage{times}
\usepackage{latexsym}
\usepackage{float}

\newcommand{\QTypes}{Question Types}

\newcommand{\Taxonomy}{Taxonomy Content}
\newcommand{\Diverse}{Diverse Content}

\usepackage{graphicx}
\usepackage{subcaption}
\usepackage{caption}
\usepackage{multirow}
\usepackage{array}
\usepackage{amsmath}
\usepackage{todonotes}
\usepackage{diagbox}
\usepackage{bm}
\usepackage{bm,bbm}

\usepackage{booktabs}
\usepackage[nameinlink,noabbrev,capitalise,sort]{cleveref}
\usepackage{amsmath,amssymb}
\definecolor{SoftGreen}{HTML}{39ac73}
\usepackage{enumitem}

\usepackage[T1]{fontenc}

\usepackage[utf8]{inputenc}

\usepackage{microtype}

\usepackage{inconsolata}

\usepackage{graphicx}

\title{Enhancing Continual Learning in Visual Question Answering\\with Modality-Aware Feature Distillation}

\author{
    Malvina Nikandrou$^{1}$
    {\bf Georgios Pantazopoulos$^{1,2}$}
    {\bf Ioannis Konstas$^{1,2}$}
    {\bf Alessandro Suglia$^{1,2}$}
      \AND \textnormal{$^1$Heriot-Watt University; $^2$Alana AI}\\
    \AND \textnormal {\texttt{\{mn2002, gmp2000, i.konstas, a.suglia\}}\texttt{@hw.ac.uk}} \\}

\begin{document}
\maketitle
\begin{abstract}
Continual learning focuses on incrementally training a model on a sequence of tasks with the aim of learning new tasks while minimizing performance drop on previous tasks.
Existing approaches at the intersection of Continual Learning and Visual Question Answering (VQA) do not study how the multimodal nature of the input affects the learning dynamics of a model. In this paper, we demonstrate that each modality evolves at different rates across a continuum of tasks and that this behavior occurs in established encoder-only models as well as modern recipes for developing Vision \& Language (VL) models. Motivated by this observation, we propose a modality-aware feature distillation (MAFED) approach which outperforms existing baselines across models of varying scale in three multimodal continual learning settings. Furthermore, we provide ablations showcasing that modality-aware distillation complements experience replay. Overall, our results emphasize the importance of addressing modality-specific dynamics to prevent forgetting in multimodal continual learning.

\end{abstract}

\section{Introduction}
Large Language Models (LLMs)~\cite{touvron2023llama, jiang2023mistral} and Visual Language Models (VLMs) \citep{bai2023qwenvl, liu2024visual}, have achieved unprecedented performance and have become the go-to option for most NLP and Vision \& Language (VL) tasks.
However, once they have been trained, it is not straightforward how to update them to accommodate for novel concepts or concepts with reworked meanings.
As a result, over time, the knowledge of these models may be obsolete or needs to be refined periodically to maintain their relevance.
Consider two examples: 1) As of July 2023, Twitter has been re-branded to X with a new logo. 
2) According to the Oxford English Dictionary (OED), as of March 2024, more than 1000 English words or phrases have been either been revised or included as novel entries\footnote{ \href{https://www.oed.com/information/updates/march-2024/}{OED March 2024 update} lists entries appearing for the first time, or entries with updated meanings.}.
In these cases, models trained with data from a preceding period will inevitably show performance deterioration \cite{lazaridou2021pitfalls}.
Commercialized LLMs circumvent this limitation~\cite{OpenAI2022IntroducingChatGPT, team2023gemini, Claude} with statements regarding the knowledge cutoff of these models.
On the other hand, humans continuously update their knowledge and acquire new skills over time.
Continual learning is a paradigm that aims to simulate this behavior, focusing on models that learn incrementally from a sequence of tasks with minimal catastrophic forgetting~\cite{mccloskey1989catastrophic,ratcliff1990connectionist}.

\begin{figure}[tb]
    \centering
    \includegraphics[width=\columnwidth]{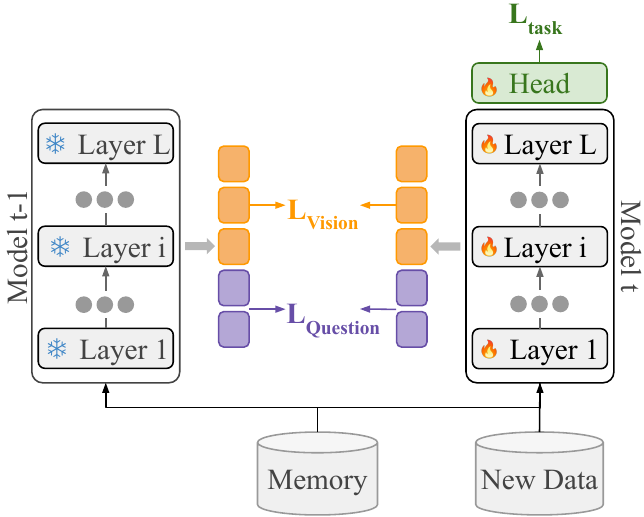}
    \caption{Overview of MAFED. Along with training on the data from the current task and a memory buffer, we apply feature distillation using the previous checkpoint as the teacher. 
    The distillation losses applied to the representations from question and visual tokens are weighted separately to compensate for modality-specific training dynamics.
    }
    \label{fig:featdistill_method}
\end{figure}

Continual learning has started being explored more widely in VL settings~\cite{greco2019psycholinguistics, NEURIPS2022_bd361197, nikandrou2022task, zhang2023vqacl, lei2023symbolic, cui2024continual}. 
However, existing approaches do not explicitly account for the dissimilarities in the representation space of multimodal inputs and their effect on the learning dynamics, which we argue is necessary for effective continual learning of VL tasks.
Previous work on the optimization dynamics of multimodal learning has demonstrated that different modalities are learned at different speeds \cite{wang2020makes, wu2022characterizing}.
Using encoder and decoder-only VLMs, we empirically showcase a similar forgetting discrepancy (\cref{sec:motivation}), which indicates that representations from each modality evolve at different rates across a sequence of tasks.

Motivated by this observation, we propose \textit{MAFED}, a \underline{M}odality-\underline{A}ware \underline{FE}ature \underline{D}istillation approach summarized in \cref{fig:featdistill_method}.
We explore different strategies for weighting the distillation losses derived from the tokens of each modality, using either fixed balanced weights or adaptive weights derived from the loss gradients computed with respect to the inputs.
We combine both variants with experience replay and show promising results across all VLM families compared to established continual learning methods.
Additionally, we conduct experiments with decoder-only VLMs ranging from 100M to 1B parameters, showing that although scale alleviates forgetting, certain settings remain challenging.
Our ablations comparing experience replay and feature distillation approach showcase that these methods are complementary and yield greater performance when combined.
Overall, our results emphasize the need to address modality-specific dynamics to effectively mitigate forgetting in multimodal continual learning.

\begin{figure*}[tb]
    \centering
    \includegraphics[width=\linewidth]{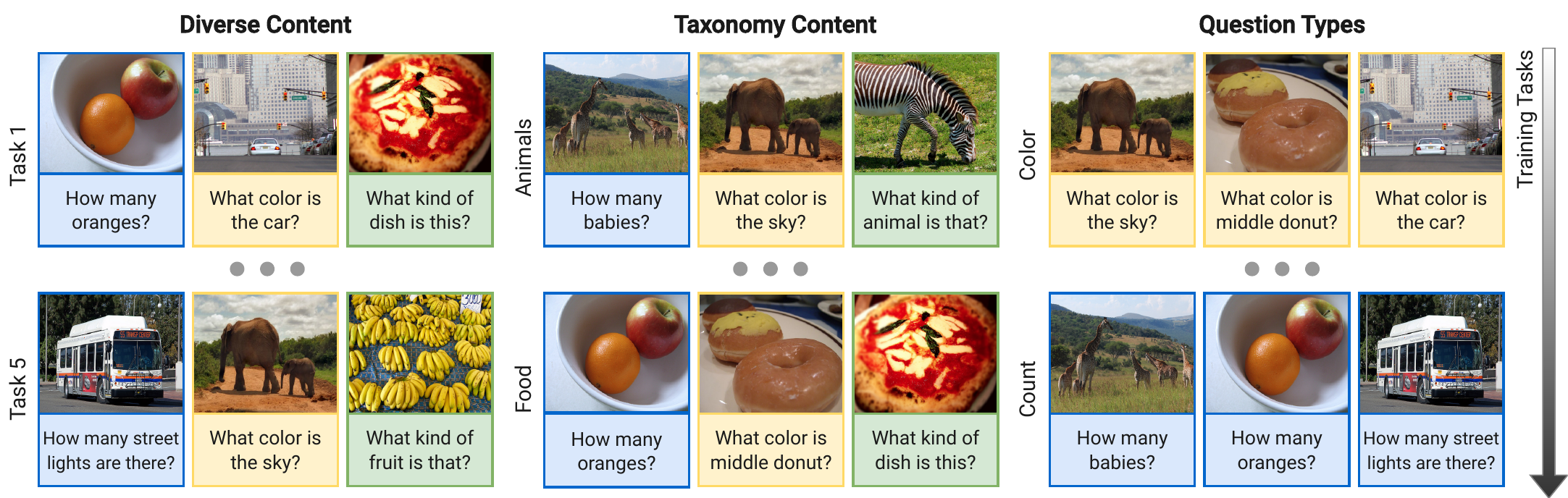}
    \caption{Illustration of tasks in each of the three continual learning settings for VQA. Each of these settings consists of five tasks. The first two settings are defined based on the visual categories. In \Diverse{}, the objects present in each task are grouped randomly, while in \Taxonomy{}, the objects are grouped based on their supercategory. Finally, in \QTypes{}, the tasks are defined according to the type of the questions.}
    \label{fig:settings}
\end{figure*}

\section{Related Work}

\subsection{Continual Learning}
\paragraph{Continual Learning Approaches}
Continual learning approaches can be categorized as regularization, replay, and architecture-based \cite{Delangesurvey}.
Regularization-based approaches introduce auxiliary losses that aim to constrain the model weights \cite{kirkpatrick2017elastic, zenke2017continual, aljundi2018mas}, outputs \cite{li2017lwf, rebuffi2017icarl}, or internal representations \cite{Hou_2019_CVPR}.
Replay-based approaches rely on storing \cite{Chaudhry2019er, der, bagus2021investigation} or generating samples \cite{van2018generative, Sun2019LAMOLLM} from past tasks so that they can be sampled along with new samples during training.
Finally, architecture-based approaches introduce task-specific parameters, either by masking the model parameters \cite{Yoon2020Scalable, DBLP:HAT} or adding new ones for each task \cite{pathnet, madotto2020adapter}.

Most recent work tends to combine techniques from multiple categories in order to maximize performance.
Similarly, our work utilizes replay and regularization through feature distillation.
Distillation has been used in various continual learning approaches.
Some focus on knowledge distillation on the output level, using the logits \cite{li2017lwf} or pseudo-labels from a past checkpoint \cite{wang2022continual, Karim_2022_CVPR}.
Other work applies distillation on the internal model representations \cite{dhar2019learning, douillard2020podnet, kang2022class}.
MAFED expands this line of work by introducing different weighting schemes to balance the distillation loss from visual and textual representations.

\paragraph{VL Continual Learning}

Previous work has studied varying instantiations of VL continual learning problems, including image captioning 
\cite{del2020ratt, nguyen2019contcap}, compositional phrase generalization \cite{jin2020viscol}, and 
task-incremental learning \cite{NEURIPS2022_bd361197}.
Within VQA, prior work has investigated continual learning based on question types~\cite{greco2019psycholinguistics}, across varying domains \cite{zhang2022cl, 10.1145/3581783.3612121}, and from a compositionality perspective~\cite{zhang2023vqacl}. 
~\citet{lei2023symbolic, nikandrou2022task} further study how VQA models evolve in different settings, including novel visual scenes or different question types.
However, these works have not investigated the effect of modality-aware methods with the exception of \citet{qian2023decouple} that focus on multimodal prompt learning for vision, text and fusion modules.
In contrast, feature distillation does not assume separate modality-specific and multimodal parameters and can be applied to more varied VLM architectures.

\subsection{VLMs}
Progress in representation learning has led to models that achieve impressive performance across multimodal benchmarks~\cite{vqav2, hudson2019gqa, li2023evaluating}.
Early approaches relied on complicated architectures~\cite{tan2019lxmert, lu2019vilbert}, and multiple learning objectives~\cite{chen2020uniter, li2021align, jia2021scaling}. 
More recently, given the rapid development of increasingly capable LLMs~\cite{touvron2023llama, jiang2023mistral, bai2023qwenlm}, these approaches have been superseded by a new paradigm where representations from visual experts~\cite{radford2021learning, oquab2024dinov} are treated as input tokens for the LLM.
This shift has led the development of modern VLMs~\cite{liu2024visual, dai2024instructblip, laurenccon2024obelics} that are based on the same underlying principles with deviations regarding the choice of the experts, or how the patch tokens are integrated into the language model.
Our experiments demonstrate the effectiveness of our approach in both encoder-only~\cite{chen2020uniter, pmlr-v139-kim21k-vilt} as well as decoder-only models.

\begin{figure*}[tb]
\includegraphics[width=\textwidth]{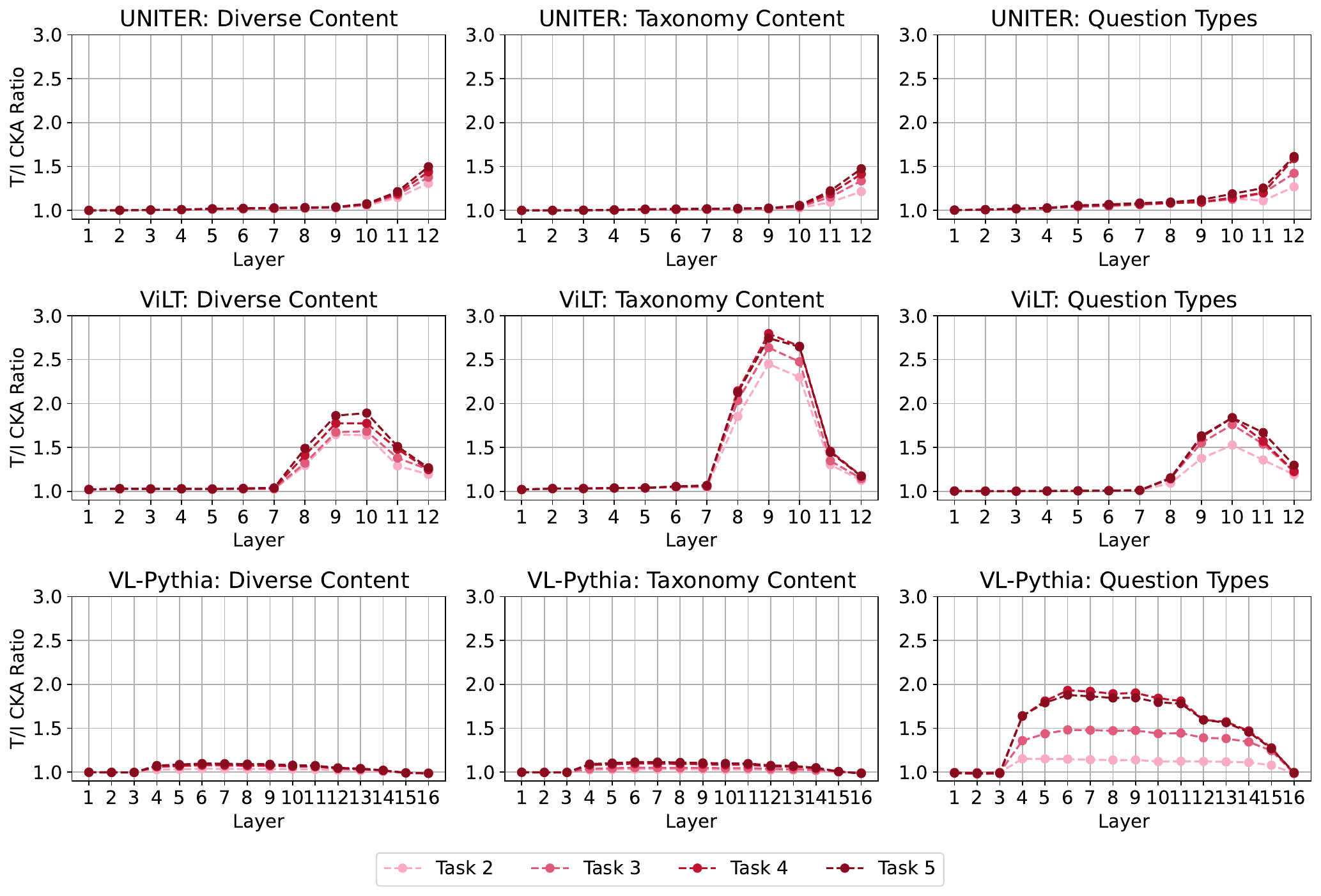}
    \caption{\small Ratio of text-to-image representation similarity across layers and tasks, for UNITER (first row), ViLT (second-row), and  VL-Pythia (third-row).
    We consistently observe that in the earlier layers, the ratio is close to one, indicating that representations from both modalities change at a similar rate. However, in intermediate or deeper layers, text representations seem to retain larger similarities.}
     \label{fig:cka_image_text}
\end{figure*}

\section{Preliminaries}
\subsection{Data}
We leverage an existing evaluation suite for Continual Learning in VQA~\cite{nikandrou2022task} comprised of three settings based on the visual and the language input.
In particular, each of these settings consists of five tasks and is designed to test the model's performance on learning varying concepts or question types. 
\cref{fig:settings} illustrates exemplary images and questions from different tasks in each of the settings.
Below, we provide a brief summary for each of these settings. 

\paragraph{\Diverse{}} corresponds to a real-world use case as well as a common standard procedure within continual learning~\cite{lomonaco2017core50,rebuffi2017icarl, zenke2017continual, lin2021the}, where a model is trained on new sets of concepts progressively that do not necessarily comply to a taxonomy.
Each task in this setting covers 10 distinct object categories from the COCO dateset~\cite{lin2014coco}.

\paragraph{\Taxonomy{}} In this setting, each task consists of questions regarding objects based on the same super-category. 
This setting contains questions from the following categories: Animals, Food, Interior, Sports, and Transport, and simulates a more progressive approach, where a model learns about a fixed (and similar) pool of concepts before being applied to a different domain. 
Importantly, we note that in both \Diverse{} and \Taxonomy{}, images containing objects shared between tasks are discarded to create clean task splits, preventing contamination between them.
In total, there are 181K train, 45K validation, and 110K test samples for both settings.

\paragraph{\QTypes{} } The final setting resembles a scenario where the model learns to answer different categories of questions. 
In this setting, the model is tasked with learning from a sequence of five tasks: Count, Color, Scene-level, Subcategory, and Action recognition.
\QTypes{} have a total of 140K train, 35K validation, and 84K test samples.

\subsection{Models}\label{ssec:models}
Throughout our experiments, we use two families of models, including encoder- and decoder-only pretrained models. 
Specifically, we use the encoder-only models, UNITER-base~\cite{chen2020uniter} and ViLT-base~\cite{pmlr-v139-kim21k-vilt}, that differ in terms of how the visual input is encoded.
UNITER uses region features extracted from the Faster R-CNN object detector~\cite{Anderson2017up-down}.
On the other hand, ViLT is a patch-based model that does not use an additional vision encoder.

However, recent trends in the development of VLMs have moved towards decoder-only architectures that combine an expert vision encoder with an LLM using a connector module that learns a mapping between them~\cite{liu2024visual}.
We employ a similar recipe to combine EVA02~\cite{fang2023eva} as the visual encoder and Pythia \citep{biderman2023pythia} as our LLM.
Regarding the connection module, we follow the LLaVA-1.5~\cite{liu2023improved} approach with a two-layer MLP that matches the dimensionality of the visual and the language embeddings.
We refer to this model as VL-Pythia.

We refrain from using other existing VLMs for two reasons. First, we aim to match the parameters and pretraining data between the encoder-only and the generative models. 
Pythia models provide a collection of checkpoints covering a wide range of sizes that have been pretrained following a state-of-art transformer recipe~\cite{su2024roformer, dao2022flashattention}. 
For a controlled comparison with encoder-only models, we train VL-Pythia using the same data used to train UNITER and ViLT (see \cref{app:pythia} for additional details regarding the pretraining of the model).
In particular, we use the checkpoints with 160M, 410M, and 1B parameters.
The smaller model is on par with the encoder-only ones, while the larger models allow us to explore the role of model capacity.
Secondly, existing VLMs~\cite{liu2024visual, dai2024instructblip, bai2023qwenlm, laurenccon2024obelics, laurenccon2024matters} are typically instruction-tuned on datasets that include VQA-v2~\cite{vqav2} on which the continual learning settings are based. This overlap is undesirable since we want to prevent any form of data contamination between tasks.

During continual learning, we keep the vision encoders of UNITER and VL-Pythia frozen. 
In encoder-only models, VQA is treated as a classification task.
The classification (CLS) token is passed to a classification head that gets expanded with the new answers from each task.
On the contrary, VL-Pythia is fine-tuned to generate answers autoregressively. 
In our experiments, we follow a greedy decoding strategy during inference.

\section{Method}\label{sec:FD}
\subsection{Motivation}\label{sec:motivation}
In this work, we argue that modality-specific learning dynamics, and more specifically, the different speeds at which each modality is learned \cite{wu2022characterizing} or forgotten, should be accounted for in multimodal continual learning settings.
We demonstrate this behavior by measuring the similarities of the question $Q$ and image $V$ representations from sequential model checkpoints using Centered Kernel Alignment (CKA)~\cite{kornblith2019cka}.
In particular, we extract text $Q_t$ and image $V_t$ representations for data of the first task after training on the first $t=1$ and each subsequent task $t=2\cdots T$, and we compute the CKA similarity across time. 
Finally, we visualize the ratio $R_t$ of the text over image similarities:

\begin{equation}
    R_t = \frac{CKA(Q_{1}, Q_{t})}{CKA(V_{1}, V_{t})} \ \quad\forall t=2\cdots T
\end{equation}

\cref{fig:cka_image_text} shows the Text-to-Image CKA ratio per layer.
We observe that there are differences in how the modalities evolve within the models and across settings.
First, we note that the ratio is greater or equal to one in all cases, meaning that visual tokens exhibit decreasing similarity throughout the learning process.
In UNITER, the ratio remains close to one for earlier layers and increases only for the last three layers.
ViLT and Pythia exhibit a different trend, where the ratio peaks for intermediate layers.
As a result, we hypothesize that incorporating the variability of each modality in a regularization technique can benefit continual learning strategies.
Due to parameter sharing between the two modalities, we materialize this in a modality-aware feature distillation strategy, which we elaborate on below.

\subsection{Modality-Aware Feature Distillation}\label{sec:method}
{\em Feature Distillation} (FD) is an established continual learning technique~\cite{Hou_2019_CVPR,douillard2020podnet,kang2022class} that adds a regularization loss term to prevent the drift of model representations.
Given two model checkpoints $f_{t-1}$ and $f_t$ from consecutive tasks, we extract representations $H$ from an intermediate layer and compute the feature distillation loss $l_i$ using the representations of each token $h_i$:

\begin{equation} \label{eq:distill1}
    \mathrm{L_{FD}} = \frac{1}{N}  \sum_{i=1}^{N}  l_i = \frac{1}{N}  \sum_{i=1}^{N}  \parallel h_{i, t} - h_{i, t-1}\parallel^2_2
\end{equation}

Assuming an example has Q text tokens and V visual tokens, we can rewrite \cref{eq:distill1} in terms of the average loss contributed by the language and the vision tokens, $ \mathrm{L_{FD, Q}}$ and $ \mathrm{L_{FD, V}}$ respectively:
\begin{equation}\label{eq:distill_per_modality}
    \mathrm{L_{FD, weighted}} = \alpha \cdot \mathrm{L_{FD, Q}} + (1-\alpha) \cdot \mathrm{L_{FD, V}}
\end{equation}

In the simplest case, $\alpha$ is proportional to the number of tokens available from each input modality. 
In practice, this might be suboptimal since it depends on the input tokenization strategy.
In fact, across the examined settings, the average tokenized inputs have approximately 9 question tokens and 33 or 199 visual tokens for region and patch-based image features, respectively.
Consequently, visual tokens will dominate the distillation loss of \cref{eq:distill_per_modality} potentially leading to inferior performance.

Therefore, we experiment with two modifications: i) MAFED-B which balances the losses from each modality by fixing $\alpha$ to 0.5, and ii) MAFED-A which uses an adaptive weighting approach based on modality importance inspired by~\citet{kang2022class}.
Modality importances $I_Q$ and $I_V$ are estimated using the gradient of the VQA classification loss\footnote{Or the language model head in the case of Pythia.}, with respect to the intermediate model representations $H_m$ from each modality $m$. $I_Q$ and $I_V$ are updated at the beginning of each training task using the available memory data $M_t$ as follows:
\begin{equation}
    I_m = \mathbb{E}_{(x,y)\sim M_t} \left[\parallel \nabla_{H_m} L_{cls}(f_t(x), y)\parallel^2_F \right]
\end{equation}
where $\parallel \cdot \parallel$ corresponds to the Frobenius norm.
Finally, the weight $\alpha$ is computed by normalizing the importance of the question  tokens:
\begin{equation}
    \alpha = \frac{I_Q}{I_Q + I_V}
\end{equation}


We apply feature distillation to all layers except the last since only the representation of CLS token in encoder-only and the final question token in decoder-only models is propagated to the model output's head.
Furthermore, in \cref{sec:motivation}, we showcased that the representations of deeper layers are affected more during continual learning.
Therefore, we introduce a discount factor $w_d$ that is used to weigh the contribution of the loss from each layer proportionally to its distance $d$ from the model's head: 

\begin{equation}
    w_d = \frac{\gamma^{d}}{\sum_{d=0}^{D} \gamma^{d}} \
\end{equation}

\begin{table*}[tb]
\centering
\small
\begin{tabular}{clcc|cc| cc}
\toprule 
& & \multicolumn{2}{c}{\textbf{\Diverse}} & \multicolumn{2}{c}{\textbf{\Taxonomy}} & \multicolumn{2}{c}{\textbf{\QTypes}}\\
\textbf{Model} & \textbf{Method} & \textbf{Accuracy} & \textbf{SBWT} & \textbf{Accuracy}  & \textbf{SBWT} & \textbf{Accuracy}  & \textbf{SBWT} \\ \midrule
\multirow{7}{1em}{\rotatebox[origin=c]{90}{UNITER}} & FT$^\ast$ & 64.59 \tiny{\textcolor{gray}{$\pm$ 0.56}} &  -1.93 \tiny{\textcolor{gray}{$\pm$ 0.39}} & 63.65 \tiny{\textcolor{gray}{$\pm$ 0.63}} & -3.89 \tiny{\textcolor{gray}{$\pm$ 0.53}} & 48.81 \tiny{\textcolor{gray}{$\pm$ 5.56}} & -22.43 \tiny{\textcolor{gray}{$\pm$ 7.02}} \\
& EWC$^\ast$ & 66.26 \tiny{\textcolor{gray}{$\pm$ 0.55}} &  -0.67 \tiny{\textcolor{gray}{$\pm$ 0.29}} & {\bf 67.70} \tiny{\textcolor{gray}{$\pm$ 0.29}} & -0.62 \tiny{\textcolor{gray}{$\pm$ 0.19}} & 66.77 \tiny{\textcolor{gray}{$\pm$ 3.54}} & -2.62 \tiny{\textcolor{gray}{$\pm$ 2.28}} \\
& ER$^\ast$ & 66.47  \tiny{\textcolor{gray}{$\pm$ 0.51}} &  -0.29 \tiny{\textcolor{gray}{$\pm$ 0.18}} & 66.76 \tiny{\textcolor{gray}{$\pm$ 0.16}} & -1.22 \tiny{\textcolor{gray}{$\pm$ 0.10}} &  69.01 \tiny{\textcolor{gray}{$\pm$ 0.76}} & -1.42 \tiny{\textcolor{gray}{$\pm$ 0.31}} \\
& FD & 66.67 \tiny{\textcolor{gray}{$\pm$ 0.38}} & -0.17 \tiny{\textcolor{gray}{$\pm$ 0.19}} & 66.94 \tiny{\textcolor{gray}{$\pm$ 0.23}} & -0.68 \tiny{\textcolor{gray}{$\pm$ 0.17}} & 69.53 \tiny{\textcolor{gray}{$\pm$ 0.12}}  & -1.22  \tiny{\textcolor{gray}{$\pm$ 0.51}} \\
& MAFED-B & {\bf 66.77} \tiny{\textcolor{gray}{$\pm$ 0.24}} & {\bf -0.12} \tiny{\textcolor{gray}{$\pm$ 0.17}} &  67.05 \tiny{\textcolor{gray}{$\pm$ 0.23}} & {\bf -0.57} \tiny{\textcolor{gray}{$\pm$ 0.08}} & {\bf 69.58} \tiny{\textcolor{gray}{$\pm$ 0.55}}  & -1.17 \tiny{\textcolor{gray}{$\pm$ 0.31}}  \\
& MAFED-A & 66.52 \tiny{\textcolor{gray}{$\pm$ 0.26}} & -0.23 \tiny{\textcolor{gray}{$\pm$ 0.12}} & 66.84 \tiny{\textcolor{gray}{$\pm$ 0.25}} & -0.70 \tiny{\textcolor{gray}{$\pm$ 0.45}} & 69.34 \tiny{\textcolor{gray}{$\pm$ 0.43}} & {\bf -0.94} \tiny{\textcolor{gray}{$\pm$ 0.48}} \\
& \textcolor{gray}{Multitask$^\ast$} & \textcolor{gray}{69.76 \tiny{$\pm$ 0.18}}  & \textcolor{gray}{-} & \textcolor{gray}{70.08 \tiny{$\pm$ 0.18}}  & \textcolor{gray}{-} & \textcolor{gray}{72.54 \tiny{$\pm$ 0.15}} & \textcolor{gray}{-} \\
\midrule
\multirow{7}{1em}{\rotatebox[origin=l]{90}{ViLT}} & FT$^\ast$ &  61.07 \tiny{\textcolor{gray}{$\pm$ 0.41}}  & -2.80 \tiny{\textcolor{gray}{$\pm$ 0.41}} & 61.25 \tiny{\textcolor{gray}{$\pm$ 0.50}}  & -4.09 \tiny{\textcolor{gray}{$\pm$ 0.50}} & 36.95\tiny{\textcolor{gray}{$\pm$ 11.09}}   & -32.86 \tiny{\textcolor{gray}{$\pm$ 11.09}} \\
& EWC$^\ast$ & 61.80 \tiny{\textcolor{gray}{$\pm$ 0.96}}  & -1.14 \tiny{\textcolor{gray}{$\pm$ 0.96}} & 63.69 \tiny{\textcolor{gray}{$\pm$ 0.46}}  & -0.92 \tiny{\textcolor{gray}{$\pm$ 0.46}} & 60.25 \tiny{\textcolor{gray}{$\pm$ 2.86}}  & -8.19 \tiny{\textcolor{gray}{$\pm$ 2.86}} \\
& ER$^\ast$ & 64.22 \tiny{\textcolor{gray}{$\pm$ 0.10}}  & -0.25 \tiny{\textcolor{gray}{$\pm$ 0.10}}  &  63.52 \tiny{\textcolor{gray}{$\pm$ 0.20}} & -1.46 \tiny{\textcolor{gray}{$\pm$ 0.20}} & 65.61 \tiny{\textcolor{gray}{$\pm$ 0.76}}  & -2.86 \tiny{\textcolor{gray}{$\pm$ 0.76}} \\
& FD  & 64.57 \tiny{\textcolor{gray}{$\pm$ 0.57}} & -0.51 \tiny{\textcolor{gray}{$\pm$ 0.12}} & 64.24 \tiny{\textcolor{gray}{$\pm$ 0.73}} & -1.07 \tiny{\textcolor{gray}{$\pm$ 0.46}} & 67.70 \tiny{\textcolor{gray}{$\pm$ 0.54}} & -1.98 \tiny{\textcolor{gray}{$\pm$ 0.70}} \\
& MAFED-B & 64.78 \tiny{\textcolor{gray}{$\pm$ 0.55}} & -0.34 \tiny{\textcolor{gray}{$\pm$ 0.27}} & 64.51 \tiny{\textcolor{gray}{$\pm$ 0.36}} & -1.02 \tiny{\textcolor{gray}{$\pm$ 0.17}} & {\bf 67.76}	\tiny{\textcolor{gray}{$\pm$ 0.27}} & {\bf -1.85} \tiny{\textcolor{gray}{$\pm$ 0.61}}\\
& MAFED-A & {\bf 65.00} \tiny{\textcolor{gray}{$\pm$ 0.41}} & {\bf -0.28} \tiny{\textcolor{gray}{$\pm$ 0.19}} & {\bf 64.63} \tiny{\textcolor{gray}{$\pm$ 0.37}} & {\bf-0.89} \tiny{\textcolor{gray}{$\pm$ 0.21}} & 67.67 \tiny{\textcolor{gray}{$\pm$ 0.46}} & -2.01 \tiny{\textcolor{gray}{$\pm$ 0.84}} \\
& \textcolor{gray}{Multitask$^\ast$} & \textcolor{gray}{67.51 \tiny{$\pm$ 1.94}}  & \textcolor{gray}{-} & \textcolor{gray}{67.84 \tiny{$\pm$ 3.92}}  & \textcolor{gray}{-} &  \textcolor{gray}{72.41 \tiny{$\pm$ 3.75}}  & \textcolor{gray}{-} \\
\bottomrule
\end{tabular}
\caption{UNITER and ViLT average accuracy and semantic backward transfer over five task orders. $^\ast$ results reported in \cite{nikandrou2022task}.}
\label{tab:vilt_uniter_featdistill}
\end{table*}

where $\gamma \in (0,1]$ is a hyperparameter such that lower $\gamma$  values assign more weight to deeper layers, and $\gamma = 1$ weighs the losses from all layers equally.
Unless stated otherwise, we combine feature distillation with replay since it requires no computational overhead and helps mitigate the miscalibration of the output layer, which can negatively impact performance on past tasks \cite{wu2019large}.

\section{Experiments}
\subsection{Baselines}
We compare our methods against naive {\em Fine Tuning} (FT), where a model is trained sequentially on each task.
Our continual learning baselines include {\em Elastic Weight Consolidation} (EWC)~\cite{kirkpatrick2017elastic} and {\em Experience Replay} (ER)~\cite{Chaudhry2019er}, which have been shown to perform competitively.
Finally, we report the upper bound performance of {\em Multitask} learning, where the model is trained on all tasks simultaneously. Note that this is the de facto standard for instruction-tuning in VLMs~\cite{dai2024instructblip,liu2023improved}.

\subsection{Evaluation Metrics}
We measure the performance of a model using three metrics. 
First, we report the macro-average accuracy at the end of a training sequence,  $\mathrm{A} = \frac{1}{T} \sum_{i=1}^T A_{T,i}$, where $A_{T,i}$ depicts the performance of the model on the data from task $i$ after training on final task $T$.
Additionally, we report the Semantic Backward Transfer (SBWT) ~\cite{nikandrou2022task}, which captures the impact of catastrophic forgetting, weighted by the semantic similarity of the prediction and the target:

 \begin{equation}
    \mathrm{SBWT} = \frac{1}{T-1} \sum_{i=1}^{T-1} S_{T,i}
 \end{equation}
 where $S_{T,i}$ is the average weighted accuracy difference for task $i$.

\subsection{Implementation Details}
We train all models using the Adam optimizer \cite{kingma2014adam} and a learning rate schedule that follows linear decay after a warmup for 10\% of the training steps.
The maximum learning rate is optimized through grid search separately for each setting based on the performance of the finetuning method.
For UNITER and ViLT, we set the number of epochs per task to 60 with a patience of 5.
We found that VL-Pythia models reach their peak accuracy for fewer updates, possibly because they do not use a randomly initialized classification head.
As a result, we set the maximum number of epochs to 15. 
For all replay and feature distillation runs, we keep a memory of 1000 randomly selected samples per task, ensuring that the same samples are stored across methods for the same task order.
Further details about the selected hyperparameters are listed in \cref{app:hyperparams}.

\subsection{Results}\label{sec:results}
\begin{table*}[tb]
\small
\centering
\begin{tabular}{clcc|cc|cc}
\toprule 
& & \multicolumn{2}{c}{\textbf{VL-Pythia 160M}} & \multicolumn{2}{c}{\textbf{VL-Pythia 410M}} & \multicolumn{2}{c}{\textbf{VL-Pythia 1B}} \\
\textbf{Setting} & \textbf{Method} & \textbf{Accuracy} & \textbf{SBWT} & \textbf{Accuracy}  & \textbf{SBWT} & \textbf{Accuracy} & \textbf{SBWT}\\ \midrule

\multirow{7}{1em}{\rotatebox[origin=c]{90}{\small{\QTypes}}} & FT & 25.98 \tiny{\textcolor{gray}{$\pm$ 8.23}} & -31.35 \tiny{\textcolor{gray}{$\pm$ 7.27}} & 63.20 \tiny{\textcolor{gray}{$\pm$ 2.10}} & -7.22 \tiny{\textcolor{gray}{$\pm$ 1.98}} & 65.52 \tiny{\textcolor{gray}{$\pm$ 5.60}} & -6.16 \tiny{\textcolor{gray}{$\pm$ 5.42}}\\
& EWC &  41.55 \tiny{\textcolor{gray}{$\pm$ 8.31}} & -8.58 \tiny{\textcolor{gray}{$\pm$ 6.96}} & 66.83 \tiny{\textcolor{gray}{$\pm$ 2.45}} & -3.85 \tiny{\textcolor{gray}{$\pm$ 2.48}} & 66.78  \tiny{\textcolor{gray}{$\pm$ 2.98}} & -5.30 \tiny{\textcolor{gray}{$\pm$ 2.93}}\\
& ER & 53.56 \tiny{\textcolor{gray}{$\pm$ 0.72}} & -5.20 \tiny{\textcolor{gray}{$\pm$ 1.06}} & 70.25 \tiny{\textcolor{gray}{$\pm$ 1.00}} & -0.52 \tiny{\textcolor{gray}{$\pm$ 0.67}} & 69.66 \tiny{\textcolor{gray}{$\pm$ 3.34}} &  -2.27 \tiny{\textcolor{gray}{$\pm$ 1.88}} \\
& FD & 56.19 \tiny{\textcolor{gray}{$\pm$ 1.59}} & -3.04 \tiny{\textcolor{gray}{$\pm$ 1.21}} &  70.76 \tiny{\textcolor{gray}{$\pm$ 0.50}} & -0.22 \tiny{\textcolor{gray}{$\pm$ 0.25}} &  71.85 \tiny{\textcolor{gray}{$\pm$ 1.07}} & -0.62 \tiny{\textcolor{gray}{$\pm$ 0.54}} \\
& MAFED-B & 57.53 \tiny{\textcolor{gray}{$\pm$ 0.76}} &  -2.83 \tiny{\textcolor{gray}{$\pm$ 0.42}} & 70.82\tiny{\textcolor{gray}{$\pm$ 0.38}} & -0.17 \tiny{\textcolor{gray}{$\pm$ 0.10}} & 72.19 \tiny{\textcolor{gray}{$\pm$ 1.40}} & -0.55 \tiny{\textcolor{gray}{$\pm$ 0.80}} \\
& MAFED-A & \textbf{57.65} \tiny{\textcolor{gray}{$\pm$ 0.24}} & \textbf{-2.46} \tiny{\textcolor{gray}{$\pm$ 0.29}} & \textbf{71.06} \tiny{\textcolor{gray}{$\pm$ 0.30}} & \textbf{-0.19} \tiny{\textcolor{gray}{$\pm$ 0.28}} & \textbf{72.69} \tiny{\textcolor{gray}{$\pm$ 0.12}} & \textbf{-0.10} \tiny{\textcolor{gray}{$\pm$ 0.07}}\\
& \textcolor{gray}{Multitask} &  \textcolor{gray}{65.65 \tiny{$\pm$0.14}} &  - &  \textcolor{gray}{71.96 \tiny{$\pm$ 0.15}} & - & \textcolor{gray}{73.44 \tiny{$\pm$ 0.18}} & - \\
\bottomrule
\end{tabular}
\caption{Performance of different VL-Pythia model sizes across three task orders.}
\label{tab:scaling}
\end{table*}

\subsubsection{Encoder-only Models}\label{sec:uniter_vs_vilt}
\cref{tab:vilt_uniter_featdistill} reports the results across the three settings using the encoder-only models UNITER and ViLT.
Adding the feature distillation loss improves upon the ER baseline in all settings.
Although the benefits with UNITER are moderate, as ER already achieves low forgetting, when using ViLT, FD offers substantial accuracy gains of up to 2.6 in \QTypes{}.
Comparing the feature distillation variants, modality-aware weighting (MAFED-A or MAFED-B) consistently boosts performance.
For UNITER and ViLT in \QTypes{}, equally balancing the modality losses with MAFED-B shows the best performance.
In the image-based settings with ViLT, adaptive weighting performs better.
Given the similarity ratios shown in \cref{fig:cka_image_text}, these results suggest that MAFED-B is more effective when the relative change of text and vision representation is small, while MAFED-A is more appropriate in cases of larger discrepancies.

\subsubsection{Scaling to larger decoder-only VLMs}
\begin{table}[ht!]
    \centering
\resizebox{0.4\textwidth}{!}{
    \begin{tabular}{lcc}
    \toprule
      \textbf{Setting}  & \textbf{Accuracy} & \textbf{BWT} \\\midrule
     \Diverse & 70.11 & 1.58 \\
     \Taxonomy & 69.03 & 0.44 \\
     \QTypes & 66.01 & -9.70 \\\bottomrule
    \end{tabular}
}
\caption{Average accuracy and backward transfer for finetuning VL-Pythia 1B across settings. We report the accuracy of three task orders on the validation set.}
    \label{tab:other_settings}
\end{table}
As decoder-only architectures have become more widely used, we also experiment with three model sizes of VL-Pythia (160M, 410M, 1B parameters).
In our initial results shown in \cref{tab:other_settings}, we find that larger models exhibit no forgetting in the image-based settings of Diverse and Taxonomy Content.
As a result, we focus on more challenging \QTypes{} setting.

\cref{tab:scaling} provides the results for different continual learning strategies using the VL-Pythia variants.
We observe that scaling leads to higher final accuracy and less catastrophic forgetting similar to previous work~\cite{mirzadeh2022wide, ramasesh2022effect}.
Nevertheless, even the largest explored model has a gap of almost 8\% between naive finetuning and multitask learning.
Compared to experience replay, we find that the benefit of EWC diminishes for larger models.
As in encoder-only models, the inclusion of feature distillation further improves performance, and modality-aware weighting of the distillation losses is beneficial in all cases.
For VL-Pythia, MAFED-A leads to the best performance, improving the accuracy on average by +2.6 compared to ER and +0.87 compared to FD.
As mentioned in \cref{sec:uniter_vs_vilt}, we hypothesize that adaptive modality weighting can be particularly effective where the text and visual representations change more unequally.
Overall, our results indicate that stabilizing the representations from vision and language tokens separately can improve multimodal continual learning.

\section{Analysis}

\subsection{Distillation without Replay}
\begin{table}[ht!]
    \small
    \centering
\resizebox{0.35\textwidth}{!}{
    \begin{tabular}{lcc}
    \toprule
      \textbf{Method}  & \textbf{Accuracy} & \textbf{SBWT} \\\midrule
      FT & 62.77 &  -5.27\\
      ER & 66.18  & -3.42\\
      \midrule
     FD & 72.05 & -1.00 \\
     w/o Replay & 63.53 & -4.87 \\\midrule
      MAFED-B & 72.91 & -0.16 \\
      w/o Replay & 67.66 & -3.67\\
    \midrule
      MAFED-A & 72.56 & -0.24 \\
    w/o Replay & 64.14 &  -4.09\\
     \bottomrule
    \end{tabular}
}
\caption{Ablation of feature distillation methods without replay using VL-Pythia (1B) on \QTypes.}
    \label{tab:fd_no_replay}
\end{table}

In the previous section, we applied feature distillation in conjunction with experience replay. \cref{tab:fd_no_replay} shows the performance when applying feature distillation with and without experience replay on VL-Pythia (1B) for one task order on the \QTypes{} setting.
Feature distillation improves the model performance, while standalone modality-aware methods are competitive with experience replay. 
The combination of the two approaches yields the greatest performance with no additional cost over standalone feature distillation.
Both methods require maintaining a memory of past samples that are passed through the current model. 
Therefore, applying replay puts no additional overhead on training time and memory. 

\subsection{Distillation Layer Ablation}
\begin{figure}[ht!]
    \centering
    \includegraphics[width=0.49\textwidth]{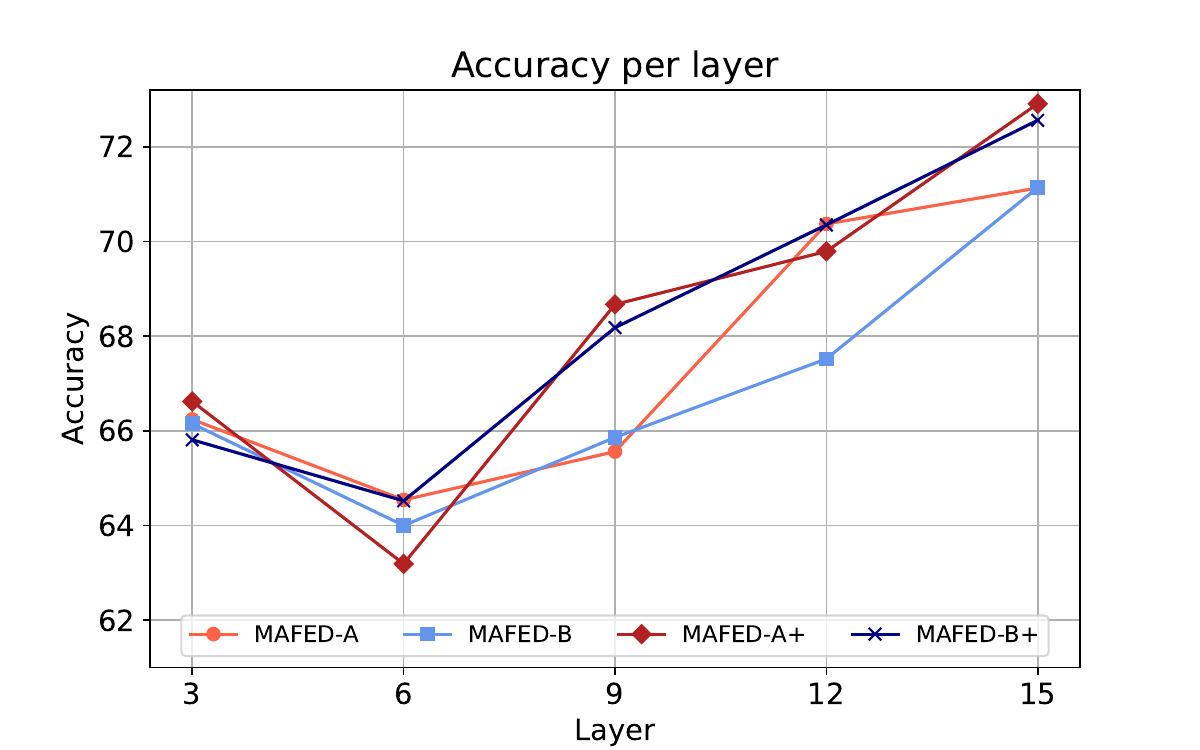}
    \caption{Ablation of feature distillation from a single or cumulative (+) model layers.}
    \label{fig:featdistill_layer}
\end{figure}
Next, we explore the effectiveness of applying modality-aware feature distillation from a single layer, as well as a subset of layers.
\cref{fig:featdistill_layer} illustrates the performance of VL-Pythia (1B) after applying MAFED-B and MAFED-A every three layers.
As expected, applying both methods on a deeper layer of the model but also distilling from all previous layers yields greater performance. 
However, the performance of all four variants does not increase monotonically with the  layer depth.
More specifically, we observe that distilling from layer 6 leads to performance degradation.
This behavior correlates with the results in \cref{fig:cka_image_text}, where the per-modality similarities diverge the most at layer 6 and gradually align throughout the deeper layers of the model.

\subsection{Modality Weights in MAFED-A}\label{sec:fda_weights}
\begin{figure}[ht!]
    \centering
    \includegraphics[width=0.45\textwidth]{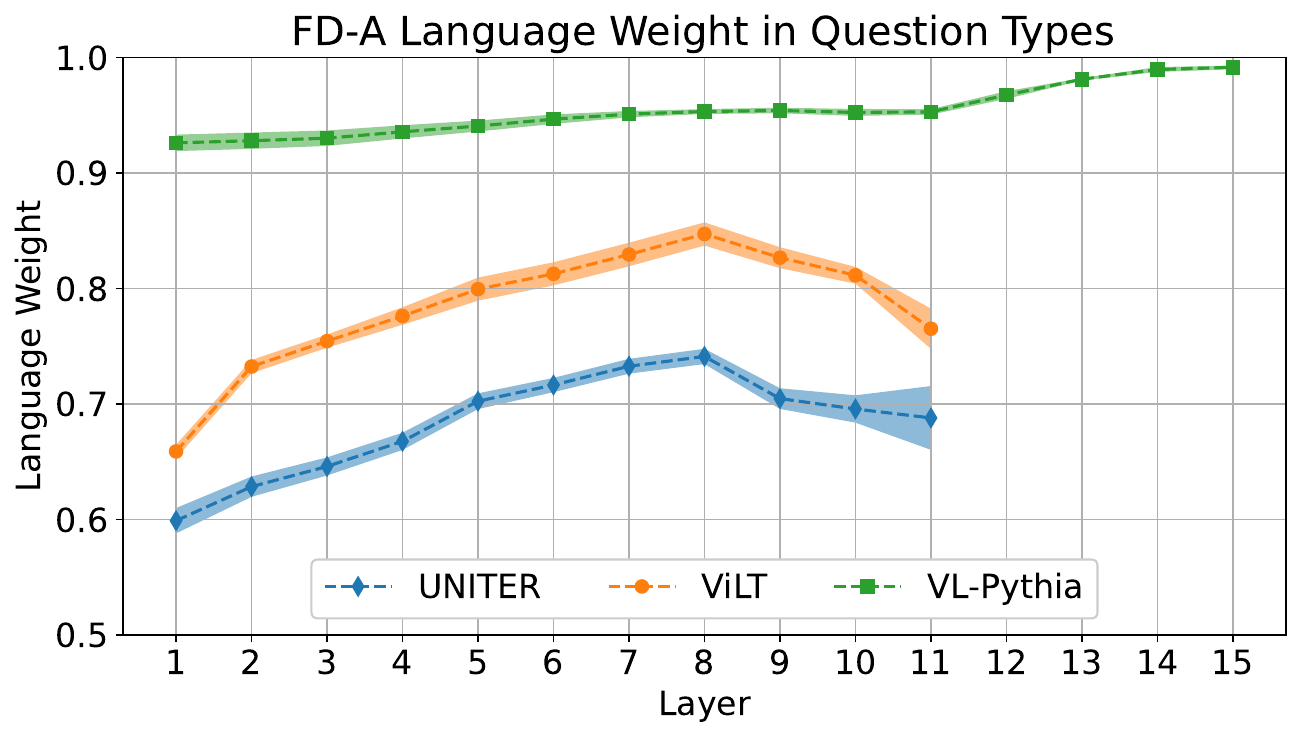}
    \caption{Language weight during MAFED-A. Note that language tokens in the encoder family (UNITER and ViLT) are weighted similarly across the layers of the models. For the causal VL-Pythia model, the language tokens have higher weights.}
    \label{fig:fda_weights}
\end{figure}

\cref{fig:fda_weights} shows the weight $\alpha$ placed on the distillation loss of the language tokens using the MAFED-A method.
In all models, MAFED-A assigns more weight to language tokens.
Interestingly, we observe that in encoder-only models, the language weight progressively increases up to layer 8 and then drops.
On the other hand, in VL-Pythia (1B), more than 90\% of the weight is assigned to language tokens for all layers.
We hypothesize that this is because the model is causal, and the last token before the answer is a text token.

\section{Conclusion}
In this paper, we argued that applying approaches that were developed with unimodal models in mind is suboptimal for continual learning in VQA since this ignores modality-specific learning dynamics. 
We empirically showcased that the visual and the textual representations evolve at different rates -- a phenomenon that occurs in both encoder-only and decoder-only VLMs.
Given this observation, we proposed two modality-aware feature distillation approaches that equally weigh the distillation loss from each modality or adaptively estimate the importance of a modality based on the gradients with respect to the inputs.
We believe this is a promising direction towards closing the gap with multitask training in multimodal continual learning.

\subsection{Limitations \& Future Work}
Despite the promising results, our method has certain limitations. 
First, distillation is more computationally expensive than replay, as it requires accessing the representations from the previous model.
However, compared to established distillation methods, MAFED-B improves performance with no overhead, while MAFED-A requires computing importance weights, which are only updated between tasks.
Furthermore, our work does not investigate the potential effectiveness of architecture-based approaches, which could offer greater control over the learning of each modality through novel parameter-isolation approaches.
Finally, we show that larger models exhibit less or even no forgetting depending on the setting.
Future work should explore whether increasing the model size or the VL pretraining data \cite{ostapenko2022continual} can further decrease forgetting in VQA settings.

\section*{Acknowledgements}
This work was supported by the Edinburgh International Data Facility (EIDF) and the Data-Driven Innovation Programme at the University of Edinburgh.

\bibliography{acl_latex}

\clearpage
\appendix

\section{Experiments}
\subsection{Pretraining VL-Pythia}\label{app:pythia}
We pretrain all VL-Pythia models following the LLaVA-1.5 recipe~\cite{liu2023improved}.
The only difference is that we skip the first stage for multimodal alignment as 
recent work~\cite{karamcheti2024prismatic} has shown that the two-stage training can be redundant and the same performance can achieved when omitting the first stage of training.
Throughout VL pretraining, the vision encoder remains frozen, while the LLM and connector parameters are trained using the Adam optimizer~\cite{kingma2014adam} with a batch size of 256 and a learning rate of 1e-3.
For all models, we used the same data to train UNITER and ViLT - COCO~\cite{lin2014coco}, SBU captions~\cite{NIPS2011_5dd9db5e}, Visual Genome captions~\cite{krishna2017visual} and Conceptual Captions 3M~\cite{sharma2018conceptual}.
We perform one epoch of pretraining and keep the final checkpoint.

\subsection{Hyperparameters}\label{app:hyperparams}
\begin{table}[ht!]
    \centering
   \resizebox{\columnwidth}{!}{
    \begin{tabular}{llcccccc}\hline
Model & \textbf{Setting} & \textbf{Batch Size} & \textbf{LR} & \textbf{EWC $\lambda$} & \textbf{FD $\gamma$}\\\hline
& \Diverse{}  & 1024 & 8e-5  & 500 & 0.8\\
UNITER  &  \Taxonomy{} &  1024 & 5e-5 & 500 & 0.8\\
&  \QTypes{}  & 512 & 5e-5 & 20K & 0.6\\ \hline
& \Diverse{}   & 1024 & 1e-5 &  500 & 1\\
ViLT  & \Taxonomy{}  & 1024 & 1e-5 & 700 & 1\\
& \QTypes{}  & 512 & 8e-5  & 10K & 0.5 \\ \hline
 & \Diverse{}  & 128 & 5e-5  & - & 0.5  \\
VL-Pythia& \Taxonomy{}  & 128 & 5e-5  & - & 0.5 \\
 & \QTypes{}  & 128 & 5e-5  & 10K & 0.5  \\

\hline

   \end{tabular}
   }
   \caption{Selected hyperparameters.}
    \label{tab:hyperparams}
\end{table}

We tune the hyperparameters using grid search based on the validation accuracy of a single task order.
For VL-Pythia variants, we use the same hyperparameters for all model sizes, as we find them to perform reasonably well.
For UNITER and ViLT, we keep the batch size, learning rate (LR), and EWC loss weight $\lambda$ reported in prior work \cite{nikandrou2022task}.
For the remaining values, we perform the following grid search:
lr $\in \{1e-5, 5.e-5, 8e-5, 1e-4\}$, EWC $\lambda \in \{500, 1K, 5K, 10K\}$, FD discount factor $\gamma \in [0.3, 1.0]$ with a step of 0.1.

\end{document}